\documentclass{article}

\usepackage[final,nonatbib]{nips_2016}

\usepackage[utf8]{inputenc} 
\usepackage[T1]{fontenc}    
\usepackage{hyperref}       
\usepackage{url}            
\usepackage{booktabs}       
\usepackage{amsfonts}       
\usepackage{nicefrac}       
\usepackage{microtype}      
\usepackage{graphicx}
\usepackage{amsmath}

\title{Episodic memory for continual model learning}

\author{{\large \bf David G. Nagy\textsuperscript{1,2}, Gergo Orban \textsuperscript{1}} \\
  \{nagy.g.david, orban.gergo\}@wigner.mta.hu \\
  \textsuperscript{1}Computational Systems Neuroscience Lab, Wigner Research Centre for Physics, Budapest, Hungary \\
  \textsuperscript{2}Institute of Physics, Eotvos Lorand University, Budapest, Hungary}

\begin{document}

\maketitle

\begin{abstract}
Both the human brain and artificial learning agents operating in real-world or comparably complex environments are faced with the challenge of online model selection. In principle this challenge can be overcome: hierarchical Bayesian inference provides a principled method for model selection and it converges on the same posterior for both off-line (i.e. batch) and online learning. However, maintaining a parameter posterior for each model in parallel has in general an even higher memory cost than storing the entire data set and is consequently clearly unfeasible. Alternatively, maintaining only a limited set of models in memory could limit memory requirements. However, sufficient statistics for one model will usually be insufficient for fitting a different kind of model, meaning that the agent loses information with each model change. We propose that episodic memory can circumvent the challenge of limited memory-capacity online model selection by retaining a selected subset of data points. We design a method to compute the quantities necessary for model selection even when the data is discarded and only statistics of one (or few) learnt models are available. We demonstrate on a simple model that a limited-sized episodic memory buffer, when the content is optimised to retain data with statistics not matching the current representation, can resolve the fundamental challenge of online model selection.
\end{abstract}

\section{Introduction}
\vspace{-0.2cm}

In a complex, structured environment that is capable of providing a practically infinite variety of possible experiences, storing them in all their detail would take a prohibitive amount of memory and would be useless in responding to novel situations. It is more beneficial for a learning agent to extract the structure of the world into a concise model -- enabling both compression and generalisation -- and store this model instead of the observations. Such a model is termed semantic memory in the cognitive science literature. At this point the question arises: what is the benefit of devoting precious resources to encoding inconsequential contingencies by storing the observed data itself, that is, what use is episodic memory?

We argue that online learning in open-ended hypothesis spaces under realistic resource constraints presents a computational challenge that makes such a memory system necessary. In an online learning scenario, observations arrive sequentially and predictions have to be continuously updated. Iterative updates of a particular model's parameters do not require storing the data, since it is sufficient to retain only the information relevant to the specification of the parameters. However, if the structural form of the model is a priori unknown \cite{Kemp2008}, then only a subset of candidate models can be tracked at any given time, since the memory cost of retaining even such compressed statistics becomes prohibitive for an infinite set of models. The inevitable information loss resulting from this restriction presents the learning agent with a delicate problem: relevance judgements, that is, decisions about what to forget and what to remember can only be based on the currently tracked models, but the initial guess for which models these should be is likely to be wrong. The reason for being wrong is that the initial data will only warrant an overly simple model or it might be misleading about the correct structure and form. The consequence of introducing such a bias in the interpretation of new experiences towards wrong models is that statistical power required for model update cannot accumulate, since the information needed for fitting those models will often be deemed irrelevant and discarded, preventing the discovery of the correct representation. 

We propose that an episodic memory can alleviate the fundamental problem of online learning described above while still satisfying capacity constraints, by retaining a \emph{selected} small subset of samples. This mini-batch allows evidence for a novel model to accumulate by retaining the contingent details of observations irrespective of how relevant they appear under the current model. We also argue that in order to take full advantage of episodic memory, its contents should be chosen selectively, so that the combination of episodic and semantic memories provide an efficient representation of the observations.


The problem explored here is intimately related to the efforts in machine learning to handle the problem of online Bayesian model selection in arbitrarily complex model spaces. There are numerous proposals for methods that deal with online model selection or model selection in infinite model-spaces \cite{Grosse2012,hjort2010bayesian} separately. Recently, there have been attempts to tackle both challenges at once in a similar setting, but these are concerned with a restricted hypothesis space over possible model forms, such as mixture models \cite{Sato2001,Fearnhead2004,Gomes2008}. Methods that are specific to a given model form have the potential to be vastly more efficient within their domain, but we are striving to find the principles for a general purpose computational architecture that is flexible enough to accommodate uncertainty in the structural form of the model \cite{Kemp2008}. To the best of our knowledge, such a scenario has not yet been explored.

\section{Learning paradigm}
\vspace{-0.2cm}
In this paper we aim to propose a solution to a fundamental problem of online model selection under resource constraints. Our main argument is agnostic to the choice of learning method, but we are adopting the Bayesian inference framework. Here, learning can be formalised as the continual refinement and updating of a probabilistic generative model, where information about unobservable or currently not observed variables, parameters and candidate world structures can all be expressed as probability distributions over latent variables. Formally, this entails the estimation of the posterior probability of parameters ($\theta$) in a given model and/or that of the model ($m$) itself:
\begin{eqnarray}
\textrm{P}\!\left(\theta\,|\,\mathcal{D}, m\right) &\propto&
\textrm{P}\!\left(\mathcal{D}\,|\, \theta, m\right) \mathrm{P}\!\left(\theta\,|\ m\right) \label{eq:post}
\\
\textrm{P}\!\left(m\,|\,\mathcal{D}\right) &\propto&
\textrm{P}\!\left(\mathcal{D}\,|\, m\right) \mathrm{P}\!\left(m\right) 
\end{eqnarray}
In case of a uniform prior probability distribution over alternative models, the model posterior is equivalent to the marginal likelihood (mLLH),  $P\left(\mathcal{D}\,|\, m\right)$. This function implements the automatic Occam's razor principle, which ensures that the simplest model that can account for the observed variance in the data has the highest posterior probability. While Eq.\ \ref{eq:post} provides a general recipe for adjusting the model parameters to data, learning can be formulated in two markedly different ways. i), In order to obtain a posterior at a particular time $T$, the whole data set $\mathcal{D}^T$ is evaluated according to Eq.\ \ref{eq:post}. ii), Online learning relies on a parameter posterior obtained at an earlier time point $T-1$ to provide a prior for the evaluation of the novel data point $x^T$:
\begin{equation}
\textrm{P}\!\left(\theta\,|\,\mathcal{D}^T, m\right) \propto
\textrm{P}\!\left(x^T\,|\, \theta, m\right) \mathrm{P}\!\left(\theta\,|\,\mathcal{D}^{T-1}, m\right)
\label{eq:online}
\end{equation}
While online learning has the same power as batch learning, it has the benefit that it is explicitly formulated such that the effect of the earlier data points is retained in the posterior calculated for $\mathcal{D}^{T-1}$. 

While such an approach to online learning liberates us from the need to retain the whole data set, the memory constraints mean that not all posteriors can be feasibly stored. In our treatment we formalise the memory constraints such that after the model has been updated, the observation is thrown away along with the sufficient statistic for all but the best performing model. We demonstrate the challenges introduced by these constraints and the proposed solution on a mixture of Gaussians model (MoG), which has the benefit of showing non-trivial model-learning dynamics while also providing an opportunity for analytical treatment. We use a version where model selection corresponds to determining the correct number of mixture components based solely on the data; parameter learning consists of finding the means for the components; while mixture weights and variance of mixture components are assumed to be fixed and known. In the following, we introduce a learning agent that only has access to a semantic memory and demonstrate that it has a propensity to discard the specific information that would allow it to change models; then, we show that the introduction of an episodic memory substantially mitigates this problem. As a benchmark, we also compare these learners to an ideal Bayesian learner without resource constraints.

\section{Semantic-only learner under constraints}
\vspace{-0.2cm}
The memory constraints introduce two main challenges: \emph{i)} the learner needs to assess the plausibility and \emph{ii)} approximate the right parameter settings of alternative models based solely on the sufficient statistics of the tracked model, without having access to the data. The absence of the original data leads to an uncertainty regarding the possible past data sets that could lead to the same available statistics, implying a probability distribution over possible past data sets. The learner needs a method for constructing such a distribution based solely on the posterior of the current model, since this contains all the information that it has retained. We propose that a natural approximation can be obtained by the assessment of the posterior predictive distribution, $\textrm{P}\!\left(x\,|\,\mathcal{D},m\right)$, of the tracked model. While the parameter posteriors of different models in general span very different spaces and are thus not comparable, all models provide their predictions over the same data space.

In order to avoid having to start learning from scratch upon changing models, the agent has to be able to extract the knowledge contained in previous models. This was achieved by approximating the posterior for a novel model via minimising the dissimilarity of the predictive posterior distributions of the old and the novel models:
\begin{equation}
\textrm{P}\!\left(\theta\,|\,\mathcal{D},m'\right) \approx
\underset{\textrm{P}\!\left(\theta|\mathcal{D},m'\right)}{\textrm{argmin}}\;
\textrm{KL}\left[ \textrm{P}\!\left(x\,|\,\mathcal{D},m\right) \,||\, \textrm{P}\!\left(x\,|\,\mathcal{D},m'\right) \right].
\label{eq:klmin}
\end{equation}
Calculating this KL divergence analytically is in most cases unfeasible. Inspired by \cite{Snelson2005a} we aimed for a compact approximation of the predictive posterior, but instead of achieving this by simply taking a likely set of parameter settings, we've assumed that the posterior comes from a simple parametric distribution family. We have found that a parametrisation of the posterior using a single mixture component effectively reproduces both the true posterior and the true predictive posterior distribution of the model.

The comparison of the different model structures required the approximation of the marginal likelihoods. For this, expected value of the marginal likelihood over the predictive posterior was calculated: 
\vspace{-0.1cm}
\begin{equation}
\left<
\textrm{P}\!\left(\mathcal{D^*}\,|\, m'\right)
\right>_{\mathcal{D^*}\sim \textrm{P}\!\left(\mathcal{D^*}\,|\, \mathcal{D}, m\right)},
\label{eq:mllh_appx}
\end{equation}
where $\mathcal{D^*}$ denotes fake data sets obtained from the predictive posterior distribution. In general, a single data point does not constitute adequate evidence for switching to an alternative model, since it lacks sufficient statistical power. If the present model estimate is correct then it can be integrated without information loss. However, models of differing form and complexity have different kinds of regularities that they can capture, and it is exactly the recurring appearance of features of the data that the current model is unable to represent that necessitates model change. Consequently, when a novel data point arrives which pushes the learner toward a change of model form but is insufficient in itself to force a switch, then the information loss prevents subsequent model change (Fig. 1a).

\begin{figure}[ht]
\centering
\includegraphics[width=\linewidth]{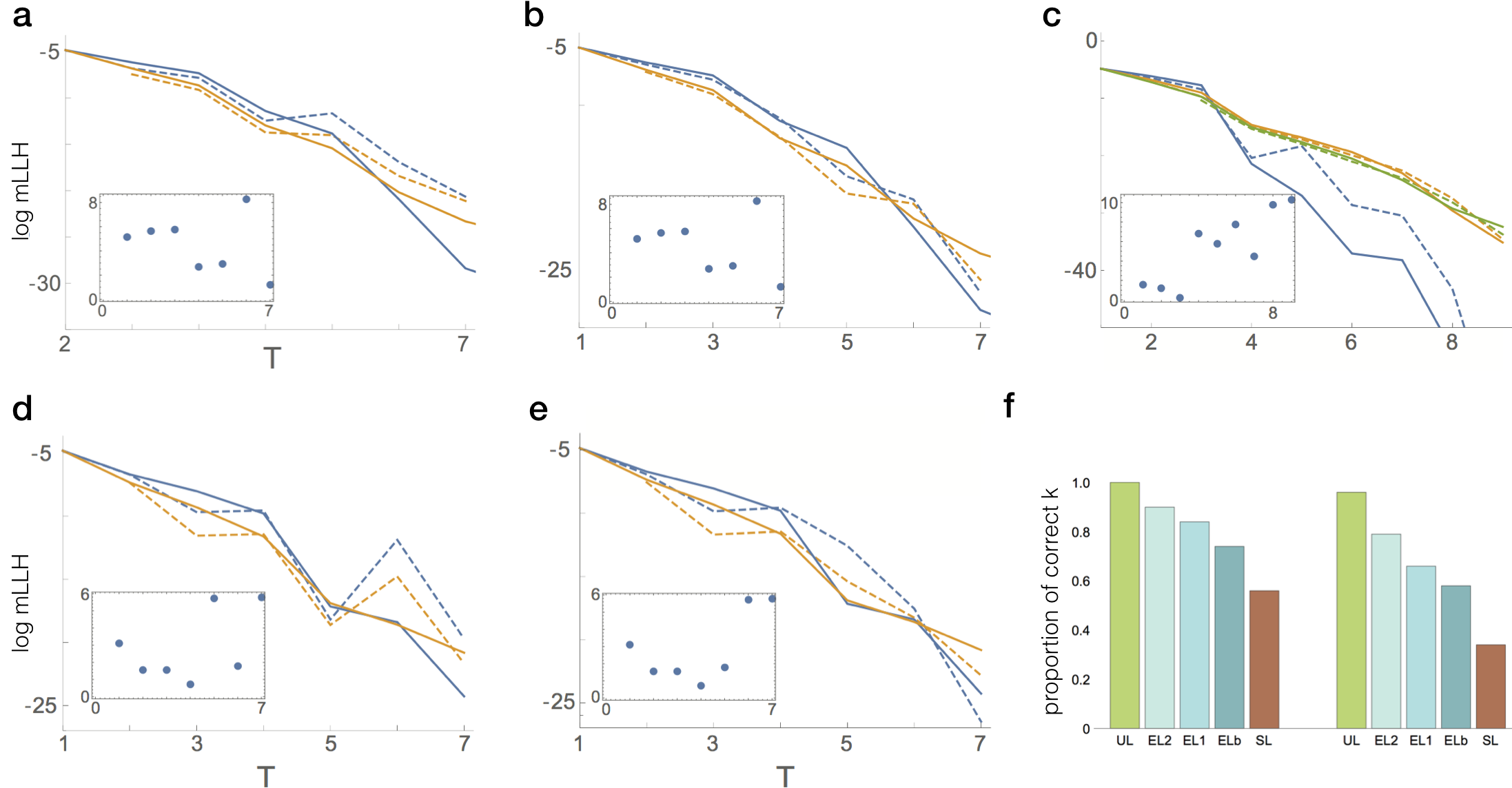}
\caption{\textbf{a}, Inability of the memory-constrained learner to increase model complexity.  Evolution of the true mLLH (analytic batch learner) of different models (\emph{continuous lines}) and the mLLH of a constrained semantic learner (\emph{dashed lines}). \emph{Inset}:the data set used. \textbf{b,c}, Effect of introducing episodic memory. Using ordered data and retaining the last two data points, the more complex model can obtain sufficient statistical power to overcome the Occam's razor effect at transitions $k=1\to2$ and  $k=2\to3$, respectively. \textbf{d}, For sampled data (unordered) a sliding window for two data points is insufficient to induce model switch. \textbf{e}, Episodic memory effectively rearranges data points such that the arrival of a subsequent data point(s) incompatible with the simple model induces model switch. \textbf{f}, Comparison of model learning in different learners. UL:unconstrained; EL2: episodic with capacity 2; EL1: episodic with capacity 1; ELb: pseudo episodic with no selectivity; SL: semantic. Probability of $k=1\rightarrow 2$ and $k=2\rightarrow 3$ model switch when data comes from a MoG with $k=2$ and $k=3$ (\emph{left} and \emph{right panels}, respectively) estimated from a thousand model runs each.
}
\label{fig:episodic}
\end{figure}

\section{Episodic learner}
\vspace{-0.2cm}
The episodic learner features an additional limited capacity storage for observations. Since the semantic learner's inability to change models is a result of loss of information about past data, providing a buffer for data points is expected to help to overcome the inherent limitation. However, we also require that the capacity of episodic memory necessary to enable model change should be small relative to the memory demands of a batch learner. Simply using an episodic storage indiscriminately as a sliding window is inefficient for enabling model change (Fig. 1d-e) since the experiences that could collectively provide the necessary statistical power for model change might not arrive consecutively. Taking full advantage of episodic storage requires the learner to optimise its contents and use it selectively.

In order to retain statistics necessary for model transitions, identifying the points that have a large information content with respect to fitting the models is paramount. We adopt the Bayesian definition of surprise \cite{itti2005}, which characterises the extent to which the posterior is different from the prior expectations: $S(\mathcal{D},m)=KL(P(m|\mathcal{D})||P(m))$. Ideally, episodes that are maximally informative regarding the model form would be sought but that would require evaluating the model posterior, $P(m|\mathcal{D}_{T-1})$, which is not accessible if the learner doesn't  evaluate the same set of models at different steps. Instead, we use the surprise in the model parameters as a proxy: this selects observations that change the learner's beliefs about the parameters the most.
%
%
%
\begin{equation}
KL(P(\mu|x_{T},\eta,k)||P(\mu|\eta,k))>\tau.
\end{equation}
Threshold $\tau$ is measured in units of surprise and a suitable value was determined empirically, but performance is relatively robust to its choice. 

We have directly contrasted the performance of learning models in the model selection task on random data sets of length $T=12$ where the generating distribution had $k=2$ or $k=3$ components (Fig. 1b-c,f). Besides the unconstrained learner and the semantic learner, we set up models for an episodic learner with a memory capacity of one and two items, and also a pseudo-episodic learner that does not perform optimisation on the items to be stored in episodic memory. The episodic learner can demonstrate a remarkable increase in performance even with an extremely limited capacity. In order to make a fair comparison between $k=1\rightarrow 2$ and $k=2\rightarrow 3$ switches we balanced the difficulty of model switch.  Our analysis on $k=2\rightarrow 3$ switch revealed an even more pronounced advantage of the episodic learner over the semantic learner, doubling the probability of a correct switch.

\section{Discussion}
\vspace{-0.2cm}
We have offered a normative argument for the existence of episodic memory by analysing a computational problem that the brain is faced with: online model selection in an open-ended model space. We used a simple minimal model to demonstrate that the introduction of memory constraints has dire consequences for a semantic-only learner and showed that these problems are substantially mitigated by an episodic memory, the contents of which are selected based on the Bayesian formalisation of surprise.

\section*{Acknowledgements}
This work was supported by a Lendület award (GO). An extended version of this paper has been presented in Nagy \& Orban (2016) \cite{Nagy2016}.
%
%


\bibliographystyle{unsrt}

\bibliography{citations}

\end{document}